# Biology-driven assessment of deep learning super-resolution imaging of the porosity network in dentin

Lauren Anderson[1,2], Lucas Chatelain[1], Nicolas Tremblay[3,4], Kathryn Grandfield[2,5], David Rousseau[6]*, Aurélien Gourrier[1]*

**Affiliations**
1. Univ. Grenoble Alpes, CNRS, LIPhy, Grenoble F-38000, France
2. McMaster University, School of Biomedical Engineering, 1280 Main St. W, L8S 4L8, Hamilton, Canada
3. CNRS, Univ. Grenoble Alpes, Grenoble-INP, Gipsa-lab, Grenoble, France
4 Department of Mathematics & Statistics, UiT the Arctic University of Norway, Tromsø, Norway
5. McMaster University, Department of Materials Science and Engineering, 1280 Main St. W, L8S 4L8, Hamilton, Canada
6. Laboratoire Angevin de Recherche en Ingénierie des Systèmes (LARIS), UMR INRAe-IRHS, Université d'Angers, 62 Avenue Notre Dame du Lac, 49000 Angers, France

Corresponding author contact :

aurelien.gourrier@univ-grenoble-alpes.fr

david.rousseau@univ-angers.fr

Abstract

The mechanosensory system of teeth is currently believed to partly rely on Odontoblast cells stimulation by fluid flow through a porosity network extending through dentin. Visualizing the smallest sub-microscopic porosity vessels therefore requires the highest achievable resolution from confocal fluorescence microscopy, the current gold standard. This considerably limits the extent of the field of view to very small sample regions. To overcome this limitation, we tested different deep learning (DL) super-resolution (SR) models to allow faster experimental acquisitions of lower resolution images and restore optimal image quality by post-processing. Three supervised 2D SR models (RCAN, pix2pix, FSRCNN) and one unsupervised (CycleGAN) were applied to a unique set of experimentally paired high- and low-resolution confocal images acquired with different sampling schemes, resulting in a pixel size increase of ×2, ×4, ×8. Model performance was quantified using a broad set of similarity and distribution-based image quality assessment (IQA) metrics, which yielded inconsistent results that mostly contradicted our visual perception. This raises the question of the relevance of such generic metrics to efficiently target the specific structure of dental porosity. To resolve this conflicting information, the generated SR images were segmented taking into account the specific scales and morphology of the porosity network and analysed by comparing connected components. Additionally, the capacity of the SR models to preserve 3D porosity connectivity throughout the confocal image stacks was evaluated using graph analysis. This biology-driven assessment allowed a far better mechanistic interpretation of SR performance, highlighting differences in model sensitivity to weak intensity features and the impact of non-linearity in image generation, which at least partly explain the failure of standard IQA metrics. Overall, CycleGAN and pix2pix were found to perform better than other models up to ×8, including for the preservation of 3D connectivity, allowing an increase of 20.3 in scan time, which paves the way to large-scale imaging required for dental research.

## 1. Introduction

Dentin porosity has attracted renewed attention following the demonstration that it may contribute to the biomechanical integrity of teeth [1]. However, since dentinal tubules that host the main cellular processes are in a range of 1-3 µm in diameter, dental porosity is currently beyond the reach of any clinical imaging device. Confocal laser scanning microscopy of *ex vivo* fluorescently stained sections thus remains the most widely used method to investigate porosity in 3D. Recently, we also showed that even smaller lateral porosity branches of 300 to 700 nm in diameter stemming from tubules act as connecting links of a vast porosity network that may also have important consequences for mechanostimulation of odontoblast cells [2], which are believed to be key players of teeth sensitivity [3]. Altogether, this is a strong incentive for a shift from standard histology analysis of dentinal porosity to connectomic approaches, as proposed for example by Weinkamer *et al.* for bone studies of the lacuno-canalicular network (LCN) [4].

The basis of such work is a faithful 3D representation of the cellular porosity network. This first implies being able to image the two biological scales of tubules and branches previously described [5]. The smallest branches are generally of the same size as the optical diffraction limit, which therefore implies using confocal microscopes with the best achievable resolution, typically ~200 nm laterally and ~600 nm axially. At such magnification, the field-of-view (FOV) is typically limited to $200 \times 200$ µm$^2$ in practice. However, in dentin, the porosity network was observed to strongly fluctuate over characteristic distances of 300 µm. Acquiring statistically relevant volumes of the porosity network therefore requires imaging teeth sections in depth over areas much larger than this 300 µm length scale, while achieving the highest possible resolution. In practice, this generally leads to prohibitive scan time.

Overcoming this challenge of significantly decreasing acquisition time without compromising resolution is generally achieved by under-sampling the biological structure and retrieving lost details by image restoration methods. Such task should advantageously benefit from recent developments in computer vision for super-resolution (SR), which aim at increasing the sharpness of low-resolution (LR) images. Note that, in our case, we can produce high-resolution (HR) data but would like to retrieve this resolution from images degraded due to faster scanning, which typically results in lower signal-to-noise ratio or under-sampling.

Many strategies exist for SR, including interpolation based, reconstruction based, learning-based, and transformer-based methods [6], [7], [8]. Deep learning (DL) approaches have shown impressive results in medical image SR and often outperform classic SR methods [8]. Most SR models are supervised and therefore rely on the availability of paired HR and LR data at the pixel level, which is somewhat experimentally challenging. Unsupervised models have therefore been developed to avoid such complex acquisitions, but ultimately, paired data are nevertheless required as ground truth (GT) to assess the image quality of restored HR data. Many studies make use of synthetically degraded images to mimic LR datasets [9], [10], [11]. However, models trained using synthetically low-resolved inputs have been heavily criticized for real-world applications [8], [12]. This defines a clear need for experimentally acquired pixel-to-pixel paired HR and LR images regardless of the DL SR model type.

The use of paired GT images to assess model performance yet provides another challenge. Image quality assessment metrics (IQA) were developed to score images based on human perception of image quality and are often used when assessing performance of DL models [13]. Most common metrics, such as peak signal to noise ratio (PSNR) [14] or structural similarity index (SSIM)[15], are used to describe how well a model performed in SR or image generation with generative adversarial networks (GANs). For example, Kim *et al.* used PSNR and SSIM to compare their SR model to state-of-the-art methods [10], Yang *et al.* compared results from their structure constrained CycleGAN model to other methods using SSIM, PSNR, and Mean Absolute Error (MAE) [16], and Zheng *et al.* used PSNR and SSIM to compare their SR CycleGAN model to other state-of-the-art methods [17]. Some studies, however, found that these common IQA metric scores did not necessarily match visual perception [18]. This raises the question whether available IQA metrics benchmarked with widely available data sets of generic scenes (e.g. landscapes, portraits etc.) are well adapted for medical or biological imaging of highly specific structures? In turn, assessing the performance of standard IQA metrics used to evaluate SR DL models, requires a specific expertise of dentinal porosity topology and connectivity analysis.

In a recent study, Jhuboo *et al* [18] performed a comparison of well-known SR models on X-ray micro-tomography (μCT) images of bone, a mineralized tissue similar to tooth dentin in composition. They pointed out two important results in our view: 1) the tested SR models didn't perform well in terms of structural indicators used for medical diagnosis and 2) classical IQAs such as PSNR and SSIM didn't necessarily correlate well to visual perception. Although the studied bone structure differs very strongly from the dentinal porosity measured in our samples, the philosophical nature of the work is similar. Reformulating their questions, we ask: how much degradation can be retrieved by a given SR model? Do models with very different architectures perform similarly? And finally, can more advanced IQA metrics than PSNR or SSIM provide better estimates of biological feature preservation in generated SR images?

The present study therefore aims to
1) Investigate the performance of four different DL SR models to restore LR images acquired through increasing degrees of under-sampling: one unpaired: CycleGAN [19] and three paired with different types of architectures: pix2pix [20], Residual Channel Attention Network (RCAN) [21], and Fast SR Convolutional Neural Network (FSRCNN) [22]. LR images are sampled by factors of x2, x4, and x8 with respect to the HR, which could decrease acquisition time by factors of 5.7, 10.2, and 20.3, respectively.
2) Assess SR model performance using standard IQA metrics by comparing generated images to matched original HR ones.

3) Compare the IQA results with a biology driven structural analysis to further assess generated image quality based on 2D morphological properties and 3D connectivity of the dentinal porosity network.

## 2. Methods

## 2.1 Tooth Slice Preparation

Two human molars were obtained following surgical extraction from 28 and 19-year-old patients, for which dentinal porosity is expected to exhibit similar porosity network characteristics and prepared using the same protocol. First, the teeth were cleaned using an ultrasound bath and disinfected with Chloramine-T solution (0.5% - 5g/L). They were then fixed for 48hrs with 70% ethanol and embedded in epoxy resin (EPOFIX, Struers). The embedded teeth were cut using a Struers diamond saw (Struers E0D15 mounted on a Presi Mecatome T210) to approximately 300 µm thick slices. One slice from each sample was selected and reduced to 200 µm thickness by lapping using 2000–4000 SiC abrasive paper (EU grade, Struers) with water lubrication and further polished using diamond sprays with grains sizes of 3 µm down to 0.2 µm (DP-spray and aqueous DP-lubricant red, Struers) using a Minitech 233 (Presi) polishing machine. The slices were rinsed with water to remove any polishing residue and left to dry for 24 hours in a petri dish and 24 extra hours at ambient room conditions. Once dried, samples were immersed in 0.02%wt Rhodamine B (83689-1G, Sigma-Aldrich) in glycerol for 72 hours to maximize porosity infiltration. The samples were then mounted in the same staining solution between a glass slide and a cover slip. To ensure maximum stability, one slice (sample A) was fixed with double sided tape to the glass slide prior to mounting and sealing was achieved using cyanoacrylate glue. Sample B was mounted without tape and sealed with nail varnish, following our standard protocol. The difference between samples is that movements at the micron scale that may occur during microscopy acquisition with our standard mounting protocol is expected to be reduced by double sided-tape fixation for sample A.

## 2.2 Confocal Fluorescent Imaging of Dentin Porosity

Images of the porosity infiltrated by the fluorescent dye were acquired using a Leica TCS SP8 confocal microscope with a 40x oil objective with 1.3NA, using a Leica immersion liquid with refractive index of 1.52. Rhodamine B was excited using a diode laser (DPSS 561) at $\lambda$ = 561nm wavelength, and the emitted fluorescent light was detected using a Photomultiplier Tube (PMT) detector with a detection range of [565 – 700 nm]. Fluorescent images therefore appear bright where there are high concentrations of fluorophore, mostly inside porosities, and dark where fluorophore has not infiltrated, inside the tissue. The most stable sample A was used to acquire the training set, consisting of 6 image stacks of 18 slices in depth near the dentin enamel junction (DEJ) for a dataset of 200 × 200 × 5.1 µm$^3$. The anatomical locations of the different scanned regions are shown in Supplementary Figure 1. Sample B was imaged in 5 regions near the DEJ shown in Supplementary Figure 2, with 19 slices per region for a dataset of 200 × 200 × 5.4 µm$^3$. The HR images were acquired with sampling steps calculated using the Nyquist criterion and theoretical point spread function (PSF) of the instrument. LR images were acquired immediately before the HR measurement, without any movement of the microscope translation stage, by changing the sampling of the image and therefore the pixel size. Full acquisition parameters are given in Table 1. For simplicity, images for each resolution will be further referenced by their pixel size increase, as labeled in Table 1.

*Table 1: Experimental imaging parameters for each image acquisition at different resolutions.*

| Image Label | HR | x2 | x4 | x8 |
| --- | --- | --- | --- | --- |
| Image Sampling (pixel$^2$) | 2048 × 2048 | 1024 × 1024 | 512 × 512 | 256 × 256 |
| Pixel Size (nm$^2$) | 100 × 100 | 200 × 200 | 400 × 400 | 800 × 800 |
| Z-step size (µm) | 0.3 | 0.3 | 0.3 | 0.3 |
| Pixel dwell time (ns) | 300 | 600 | 1200 | 2400 |
| Line Averaging | 3 | 1 | 1 | 1 |
| Scan Time (s) | 264 | 46 | 23 | 13 |
| Decrease in Scan Time (Factor) | | 5.7 | 10.2 | 20.3 |
| Ratio of Pixels : PSF | 2:1 | 1:1 | 0.5:1 | 0.25:1 |

An example of images from each resolution in one acquired region at the DEJ is shown in Figure 1, with an example of a tubule indicated by a blue arrow and a branch by a green arrow. Images at decreasing resolutions clearly show a loss of information, specifically in small branch structures, as illustrated by two branches circled in green in Figure 1.

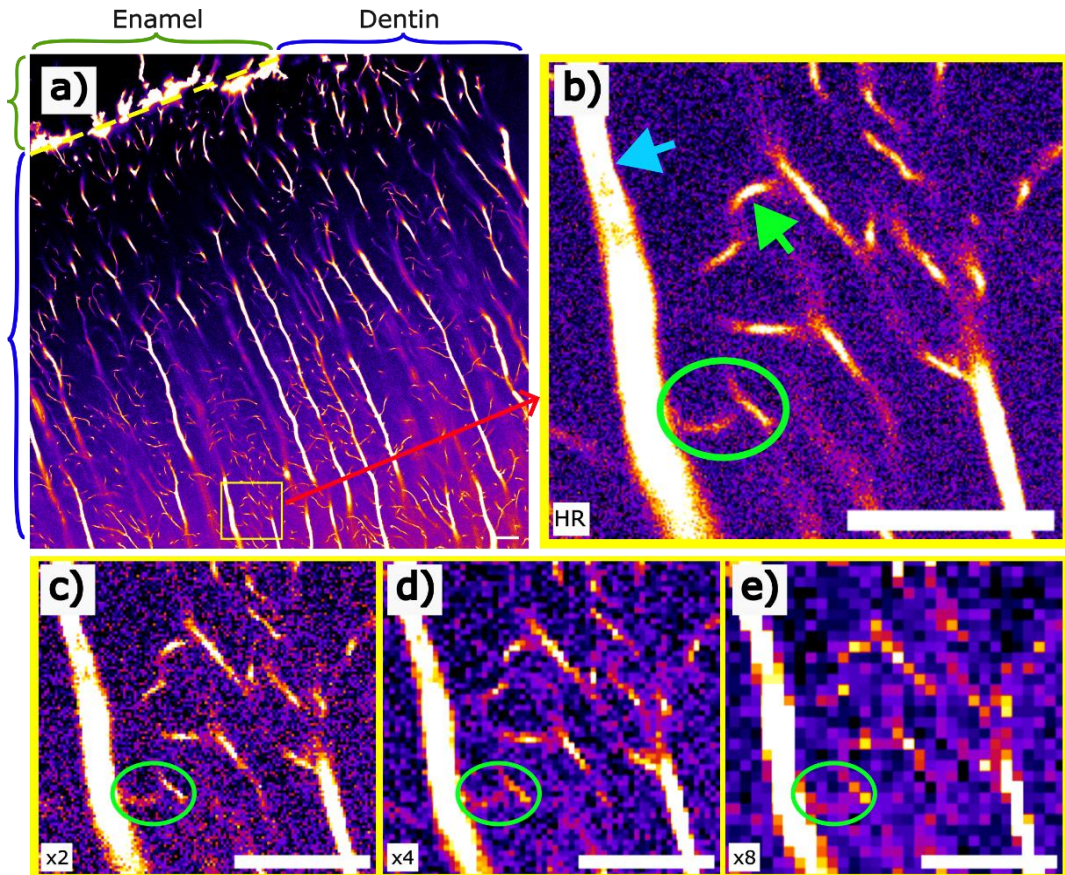

*Figure 1: Examples of a) one full FOV imaged in HR, with smaller region of interest in b) HR, c) x2, d) x4, and e) x8 acquisitions displaying decrease in resolution with increased pixel size acquisitions. Fluorescent filled porosity is shown by bright intensity pixels, while unstained background appears as dark pixels. One tubule is labelled with a blue arrow, and one branch is labelled with a green arrow. Green circles show an example of small branches whose detail is lost with decreasing resolution.*

## 2.3 Dataset preparation for deep learning model training

The overall methodology for this study is illustrated in the pipeline displayed in Figure 2. This pipeline describes all major steps of the work, which are described in detail in the following section.

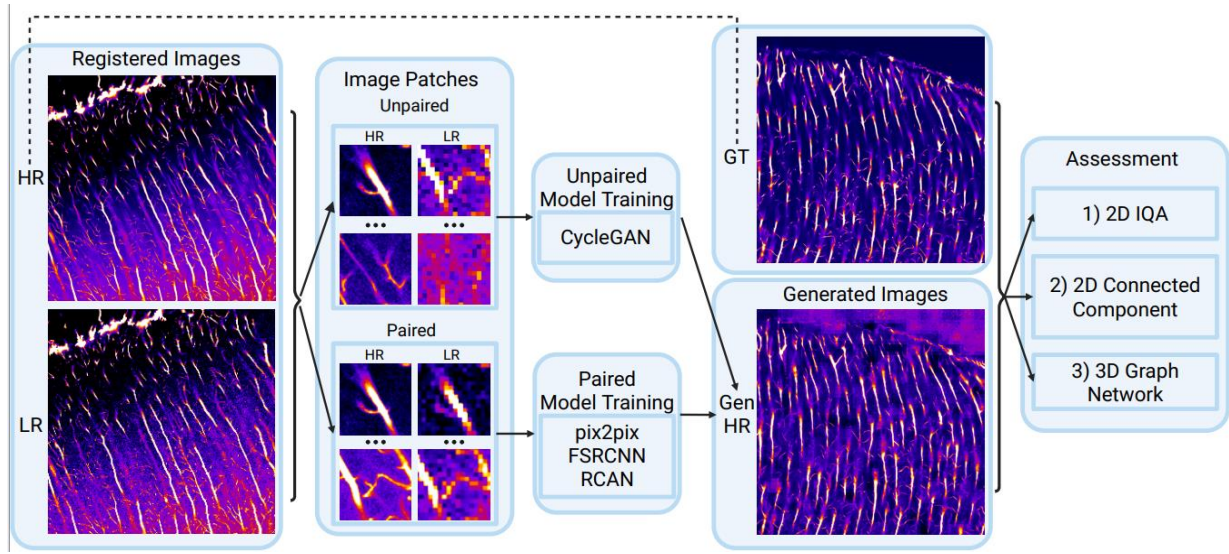

*Figure 2: Pipeline illustrating the analytical protocol. Step 1: registration of LR to HR images. Step 2: Patching whole FOV images into paired and unpaired datasets. Step 3: Training paired and unpaired models on training patches. Step 4: Applying trained models to generate HR outputs. Step 5: Assessing model performance with an image quality assessment, connected component analysis, and graph network analysis. Image created in Biorender.com.*

All data pre-processing was performed using custom code in Python 3.10. In addition to the effort undertaken to reduce experimental drift, image registration was performed to ensure fully paired data. First, all LR images were resampled to the HR format of 2048 × 2048 pixels using nearest neighbour interpolation. The python library ANTsPy [42] was then used to register LR images to the HR stack, using mutual information to find the rigid transformation between images. Registration was performed in X-Y, with no need for Z-registration due to small stack acquisitions and absence of experimental Z-drift. Images from Sample A were registered and used to create a paired training dataset. Sample B was left out to be used for testing after model training. Registered images from sample A were split into 128 × 128 pixel$^2$ patches, using 50% overlap during patching. Patches were labeled as one of 3 classes of biological features, either containing "Branches", "Tubules" or "Both". Since images were taken at the DEJ, regions containing enamel were cropped prior to image patching to solely focus on dentin and remove artefacts from the training data. Examples from each patch are shown in Figure 3.

A total of 200 patches from each class were randomly selected from each region, generating a dataset of 3600 images. The same patches were taken from the HR, x2, x4, and x8 datasets creating dataset pairs in all resolutions, and data pairs could be further shuffled to create an unpaired dataset. To augment the dataset, each patch was rotated 4 times, where each rotation angle is randomly selected from a list of angles between 0 and 180º with a 5º increment. Data augmentation using rotation was selected because in dentin, porosity can go in different directions depending on the region of dentin and how the images were acquired. Using rotation provides images with tubules and branches in all directions, making the dataset more representative and robust to different images of dentin porosity. Rotation was applied to the training and validation sets, and all images used during validation or testing were unseen in training. While this approach could cause some redundancy in the dataset, the use of overlapping patches increases the amount of data available and provides more patches to select from when creating the dataset. Multiple rotations can be applied to each patch to increase the size of the dataset and to better represent tubule orientation in different samples, thus increasing the generalizability of the dataset. While repeating structures could cause bias in model training, different tubules and branches are often quite similar in shape and size for images taken from the same regions, and therefore we can assume that the model could still generate new branches or tubules based on learning from repeated structures.

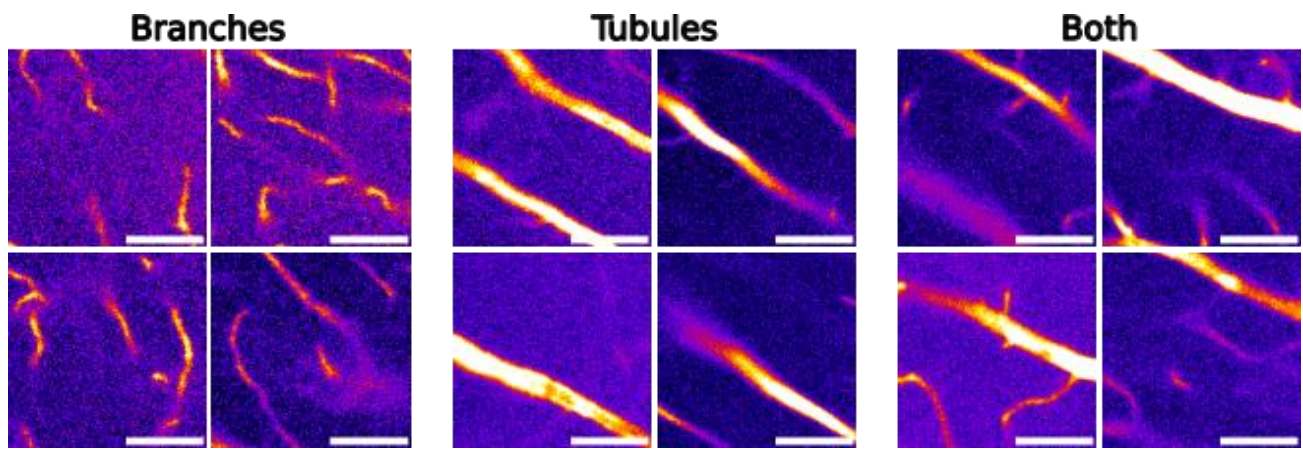

*Figure 3: 128 × 128 pixels patch showing four examples from each class: 'Branches', 'Tubules', and 'Both'. Scale = 5 µm*

The final dataset contained 18000 patches, which were separated with a 75/25 split resulting in a training set containing 13500 patches, and validation set containing 4500 patches. The datasets for x2, x4, and x8 contain the same image patches in their respective resolution, and all models were trained using the same set of patches, however for the CycleGAN, image pairs were shuffled to create an unpaired dataset.

## 2.4 Deep Learning Model Training

All SR model codes were written using python 3.10, using either tensorflow 2.10 [23] or pytorch 2.0.0 [24] based on publicly available sources: CycleGAN was modified from a keras implementation [25] with a multi-GPU implementation [26], based on [19]. Pix2pix was adapted from a tensorflow implementation [27] based on [20]. RCAN and FSRCNN implementations were derived from [18], based on [21] and [22], respectively. CycleGAN and Pix2pix model parameters were taken as was suggested by the authors and RCAN and FSRCNN hyperparameters were those used in [18]. Model parameters used for each model are described in Supplementary Table 1.

All models were trained with the previously described input data for 100 epochs. For RCAN and FSRCNN, PSNR was calculated on the evaluation set for each epoch with the intention of selecting the model with the best PSNR. However, the best PSNR did not match visual assessment. Therefore, for all models, the epoch was selected based on visual assessment of best performance over all epochs. For the CycleGAN x4 and x8, there was clear failure in model training past 40 epochs, and therefore the models at 40 epochs were used. Multiple training iterations of the CycleGAN were tried to see if different runs would produce better outcomes, but the same results were found for each run. Pix2pix, RCAN, and FSRCNN models were trained on an NVIDIA RTX A5000 GPU with 24GB RAM, and the CycleGAN model was trained using distributed training on two NVIDIA Tesla V100 GPUs with 32GB RAM. Model losses were recorded during training and can be seen in Supplementary Figure 3.

After training, models were applied to the full 2048 × 2048 pixels LR images acquired from sample B as a test set previously unseen by the models. Each image of the stacks was decomposed into a mosaic of patches using a 128 × 128 pixel$^2$ sliding window with 25% overlap. The SR models were applied to the image patches, and the generated patches were then restitched together using averaging for overlapping pixels. The python library empatches [28] was used for patching and restitching. The final mosaic images obtained after model application will further be referenced as "generated images".

After model application, it was observed that the CycleGAN model had large variation of background intensity across patches, which revealed the patching grid across the image. This problem seems to be inherent to this model. To fix this issue, the average background from each patch was estimated using the following procedure:
- the sum of pixel intensities was calculated and used to sort each patch from lowest to highest
- the first 10% of pixels (representing the lowest intensities) were assumed to be "background" due to the sparsity of the porosity features and used to calculate an average background value for each patch
- the minimum value between all average patch backgrounds was taken as a background baseline
- each patch with average background value above this baseline was then scaled to the baseline using equation 1:

$$I_n = I - (min(I) - B) \qquad (1)$$

where $I$ is the original image, $I_n$ is the new image, and $B$ is the baseline. This procedure allowed homogenizing the background intensity of the mosaic of patches across the image. An example of the effect of scaling patch intensity is shown in Supplementary Figure 4. All pixel intensities in each patch were scaled by the same value to maintain the patch contrast, keeping the integrity of model performance for subsequent evaluation and comparison to different models.

## 2.5 Trained Model Performance Assessment

Generated images were then compared to the GT dataset to assess model performance. All models were applied to the data acquired from sample B, and therefore there was no difference in quality assessment between paired or unpaired models. Five regions were compared. Images were cropped to remove enamel, register to the HR images, and have consistent image size across all regions. Image size was 1500 × 2048 × 19 pixel$^3$, 1499 × 2043 × 19 pixel$^3$, and 1499 × 2038 × 19 pixel$^3$ for x2, x4, and x8 images, respectively. More drift was observed in LR images, causing the region of unregistered data to be slightly larger as resolution decreased. Therefore, registered regions were different sizes for each resolution.

### 2.5.1 2D Image Quality Assessment Metrics

The first quality assessment was done using standard IQA metrics. With the availability of GT data, full-reference methods were used. Two different types of metrics were calculated: similarity-based and distribution based. The similarity-based metrics calculated were the SSIM [29], multi-scale SSIM (MS-SSIM) [30], information weighted SSIM (IW-SSIM) [31], PSNR [14], mean squared error (MSE), normalized cross-correlation (NCC) [32] and the Haar wavelet-based perceptual similarity index (HaarPSI) [33]. The distribution-based metrics calculated were the inception score (IS) [34], Frechet inception distance (FID) [35], kernel inception distance (KID) [36], geometry score (GS) [37], and the Wasserstein distance (WD) [38]. For similarity-based metrics, one score was found per image of the stack, and an average score was then calculated for the entire stack. For distribution-based methods, an InceptionNet v3 pre-trained on ImageNet data was used to extract features from the entire stack of images, and one score per dataset was calculated on these features [39]. One exception was the Wasserstein distance, which was calculated using the probability distribution function histogram distribution of the pixel intensities for each image and averaged across the stack. A score for all metrics was found for stacks from each region, and an average score from all regions was calculated for each model and resolution. The Python Image Quality (PIQ) library was used to calculate most metrics [13], except for NCC and WD. NCC was calculated using numpy [40], and WD was calculated using scipy [41].

### 2.5.2 Biology Driven Assessment

A connected component (CC) analysis was performed to assess the morphology of porosity features in the 2D images, and a graph network (GN) analysis was performed to assess pore connectivity of the 3D stacks.

#### Image binarization

For both approaches, a standard data processing pipeline developed for dentin porosity analysis [2] was applied to the images to binarize the images and calculate diameter maps. This pipeline was applied twice: once in 2D and another in 3D. GT and all generated images were processed identically, unless specified.

First, a vesselness enhancement filter initially proposed by Frangi [42] was used to enhance tubular structures in the image. This allows tubules and branches to be enhanced, even in low intensity cases. A vesselness enhancement filter works by identifying tubular structures based on the eigenvalues of the hessian matrix. One eigenvalue in 2D, or two eigenvalues in 3D, should be large representing a high change in intensity across a tubule section. The other eigenvalue should be small with respect to the two others, to account for relatively constant intensity along the tubule. To account for the differences in tubule diameters, vesselness filters are generally iterated at different scales. Those were selected based on a possible range of the full width half max (FWHM) across branches and tubules. The range of 2 and 24 with a step size of 0.5 was selected for 2048 × 2048 pixels images, representing 0.2-2.4 µm based on our HR scanning parameters. Jerman's implementation of the Frangi vesselness filter [43] was performed using the pyvesselns library in python [44] with a filter sensitivity, τ, set to 0.5.

Second, a hysteresis thresholding was performed to binarize the images. In 3D, images were resampled in Z by a factor of 3.5 to have isotropic voxel size. The high and low thresholds for the hysteresis filter were determined by a multi-otsu thresholding algorithm. The values found were divided by two to be more conservative of small and low intensity branches. The overall final thresholds used were 0.1 and 0.3 for our dataset. Images generated with CycleGAN showed enhanced size and intensity of porosity features, so larger thresholds were therefore set to 0.3 and 0.5. Multi-Otsu and hysteresis thresholding were performed in python using scikit-image [45].

Finally, a diameter map was calculated from the binarized images, only for the 3D images. To find diameter for each voxel, a largest fitting sphere algorithm was used. The value of each voxel is determined by the radius of the largest sphere that both fits in the foreground structure and engulfs the voxel, thus creating a diameter map. This was performed using a modified version of the porespy library [46].

## 2D Connected Component Analysis

CC analysis was performed on each stack of binarized images per region for each model using the CV2 python library [47]. Components with less than 16 pixel$^2$ area were rejected as noise. The histogram distributions of the component areas were calculated for each stack with a binning of 100. Histogram distributions are plotted in Supplementary Figure 5 and were used to calculate the Wasserstein distance between generated and GT CC maps.

For one image in each stack, components between generated and GT images were compared in detail to find several metrics. Each case is illustrated in Figure 4:

- False Positives: The number of false positives generated by the model that are not in the GT
- Missing: Components in the GT that are not in the generated image
- Merged: The number of components in the generated image that are represented by multiple components in the GT
- Split: The number of components in the GT image that are represented by multiple components in the generated image
- Matching: The number of components that are the same in both generated and GT images, without counting merged or split cases.

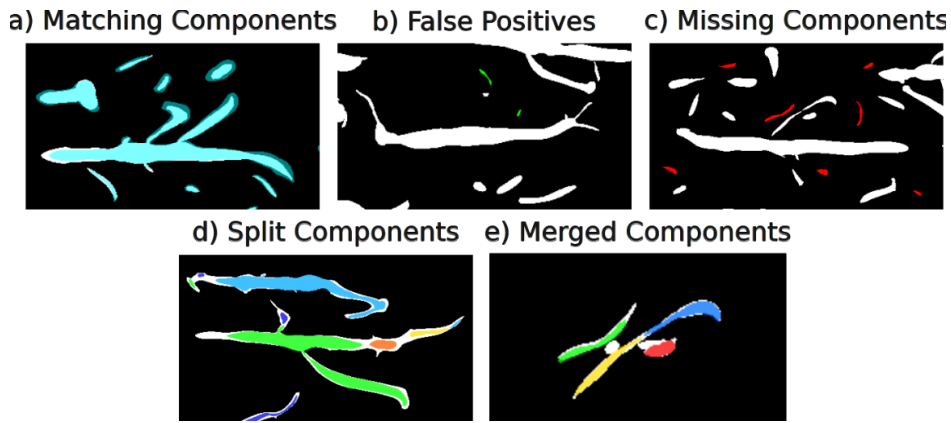

*Figure 4: Different case types for CC analysis. a) Matching component: In white, components from the GT image. In cyan, components from a generated image. b) False positives: In white, components from the GT image. In green, components from a generated image. c) Missing components: In white, components from the GT image. In red, components not found in the generated image. d) Split components: In various colours, components from the generated image. In white, components from the GT image. e) Merged components: In white, components from the generated image. In various colours, components from the GT image.*

To find all components for each case, each individual component in the generated image was checked against all components in the GT to find all cases of overlapping pixels. The same was done for each individual component in the GT compared to all components in the generated image. Only one image per stack was used due to time limitations, as comparison between each individual component, for one image from each region, model, and resolution was time consuming (approximately 5 days for the full CC analysis on Dell Precision 5820 equipped with an Intel® Xeon(R) W-2245 CPU @ 3.90GHz × 16 and 128 GB RAM).

3D Graph Network Analysis

A graph network analysis was performed on the 3D binarized and diameter map images for GT and generated images. Graph extraction protocol was previously described in a previous study [2]. Binarized images were skeletonized to reduce vessels down to their centreline. The network was then extracted from the skeleton using sknw python library [48]. Generated graphs were composed of edges and nodes, where edges represent tubules and branches and nodes represent connections or end points. Diameter mappings are also used to label edges as either branches or tubules, which is done based on diameter and direction of the structure. Manual correction was done using Fiji [49] to change any misclassified branches or tubules to their correct label. Several graph metrics were extracted from manually corrected graphs. First, the number of edges for branch, tubule and all edges. Next, the number of nodes is counted based on their degree (1, 3, 4, 5). Node degree represents the number of edges connected to each node. Finally, total edge length for branches, tubules and all edges [2].

### 2.5.3 Statistical Analysis

A statistical analysis was performed on each metrics with 95% confidence interval to assess potential differences between the 5 measured regions, showing how an average value could differ between different results from the same model. The Freidman test was used to determine if there was any significance between models and Nemenyi post hoc test was used to perform a pairwise significance test between models [50]. Three levels of significance were scored: $p < 0.05$, $p < 0.01$, and $p < 0.001$, labeled by '*', '**', and '***', respectively. Confidence intervals and p-values were calculated using the python library scikit-posthocs [51]. Statistical analysis was therefore performed for two reasons: first, to assess how confident average scores were over several results from the same model, and second, to assess how significant differences in metric scores were between models.

# 3. Results

## 3.1 Qualitative Visual Assessment Of Model Performance

Figure 5 illustrates the results of the four SR models applied to a representative image patch at the tested resolutions degradation. Qualitatively, all models seem to perform well at x2 since all tubules and branches seem to be present, although some of the later appear much fainter with respect to HR. CycleGAN maintains tubules and branches particularly well up to x4 and increases LR intensity to a similar level as HR. At x8, CycleGAN features are slightly more blurry with less defined edges, and are at a lower intensity compared to x2 and x4 results. FSRCNN rapidly fails at x4 and x8, with some tubules and branches showing much fainter at x4, and most information lost at x8. Pix2pix and RCAN perform similarly, with some lost branches across all resolutions, and all structures appearing at lower intensity than HR. Some branches were maintained but at a very low intensity making them difficult to see, as illustrated for all pix2pix resolutions by a green arrow in the bottom left corner (Figure 5). At x8, pix2pix appears to perform slightly better than RCAN, showing more defined edges for branches, however there are still many lost details compared to HR. Overall, CycleGAN seems to be at least as performant if not better than pix2pix and RCAN, appearing to maintain the most branches compared to HR, which has important practical consequences for microscopy since unpaired multi-resolution data are much easier to acquire than pixel-level paired ones. While interesting information can be deduced from visual perception of these images, qualitative assessment remains biased and therefore introduces a need for quantitative analysis of these images.

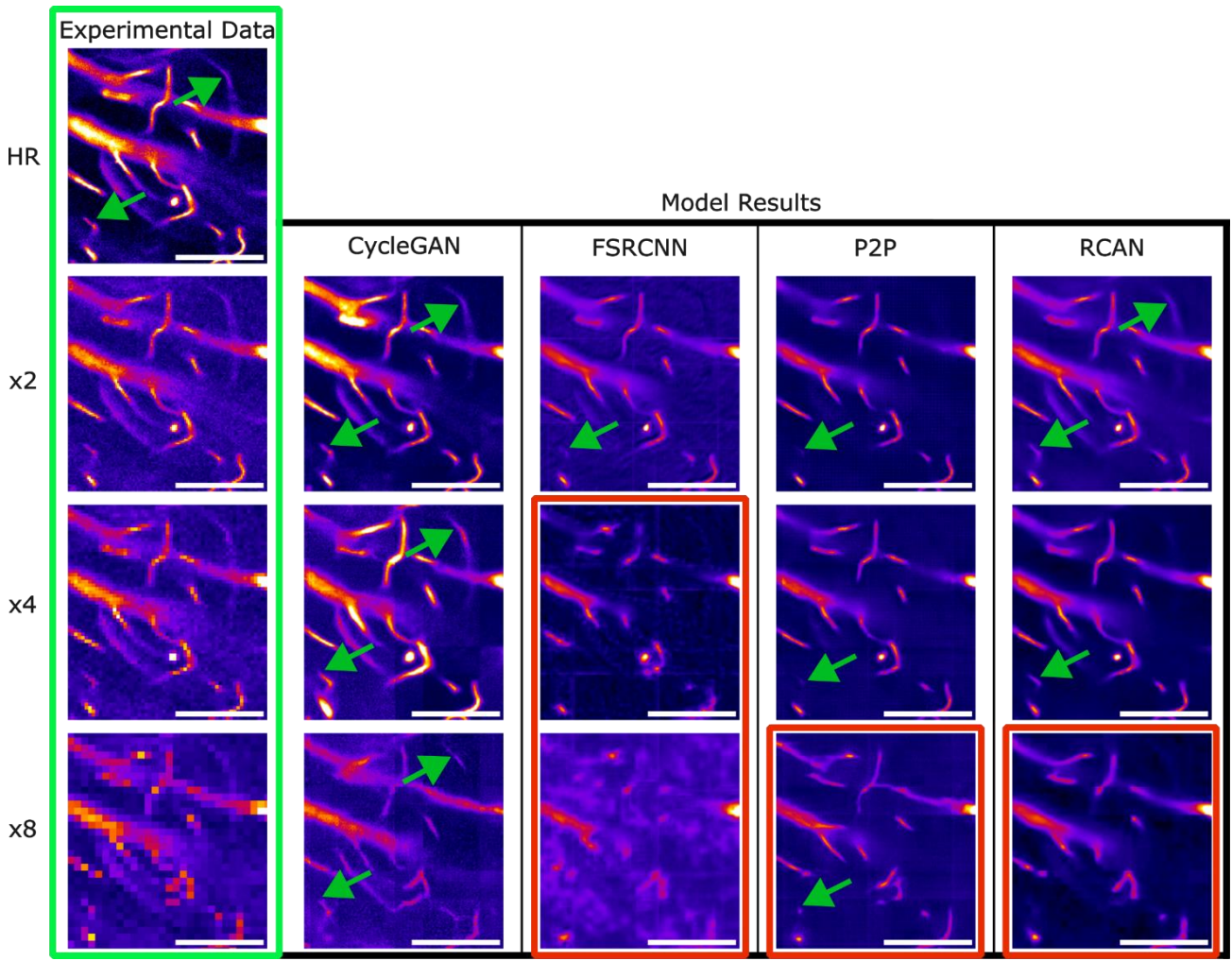

*Figure 5: SR results for a representative image patch at each experimental resolution. The GT HR and matched experimental LR image patches are shown in the first column, labelled by a green border. Green arrows highlight low-intensity branches that were maintained in some models but not others. Red borders show models which qualitatively perform poorly, with generated images far from GT. All images are displayed with the same intensity range for visual comparison. Scale = 10 µm.*

## 3.2 2D Image Quality Analysis

Figure 6 shows the results for a selected representative subset of IQA metrics. All other metrics results can be found in Supplementary Figure 6. Similarity based metrics (PSNR, SSIM, HaarPSI) clearly suggest that RCAN has the best performance, close to pix2pix and FSRCNN, which were not found to be statistically different, and CycleGAN consistently performing significantly worse compared to RCAN. PSNR decreases slightly across resolutions for all models except CycleGAN, such that the difference between models is lesser. SSIM differences are more consistent across all resolutions, remaining above 90% for RCAN and below 70% for CycleGAN. HaarPSI qualitatively behaves similar to PSNR except for pix2pix, which score remains close to RCAN, and CycleGAN, whose score is found to decrease along with other models but is always under-performing with respect to RCAN and pix2pix.

Distribution based metrics (WD, FID, GS) are globally consistent with similarity ones but exhibit more fluctuations. RCAN and pix2pix have low scores for WD and FID as high as x8 and systematically significantly outperform CycleGAN. FSRCNN is most difficult to assess: the FID puts it close to RCAN and pix2pix at x2 and worse beyond that, which is in good agreement with our qualitative assessment. But this trend is not clear with the WD, although no statistical differences were found between those three models. However, the variance between different imaged regions is considerably greater for all distribution-based measures than similarity ones, which makes it difficult to conclude with certainty. This is particularly true for the geometry score GS, which one would expect to better consider the data topology, but allows no clear interpretation in our case.

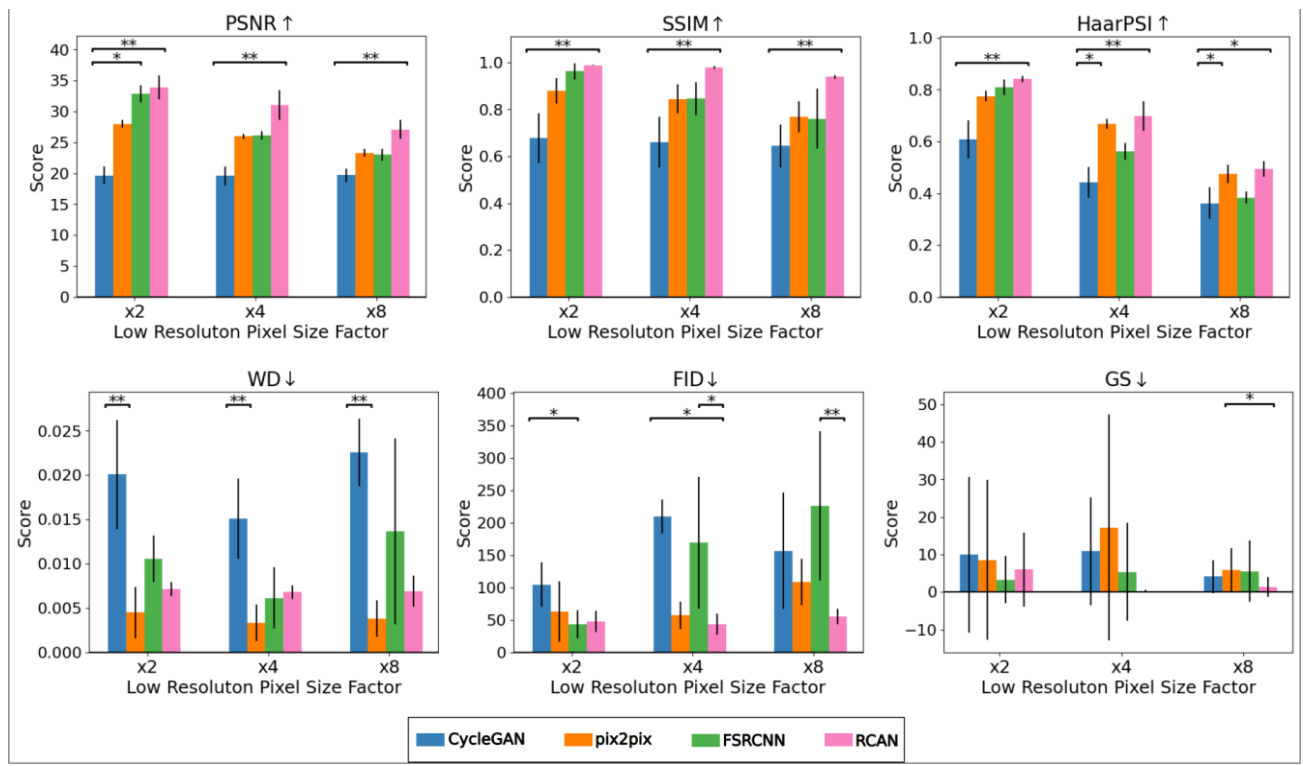

*Figure 6: IQA metric scores for all model and resolutions. Similarity based metrics PSNR, SSIM and HaarPSI are shown on the top row. Distribution based WD, FID and GS appear on the bottom row. Vertical error bars represent the variance for each score measured in the 5 imaged regions. Statistical differences between models are indicated with corresponding p-value when significant. Upward and downwards pointing arrows beside metric names detail whether high scores or low scores represent better results.*

The most important conclusion from this analysis is that IQA assessment systematically indicates CycleGAN outperformed by other models and pix2pix and RCAN as the best, which is quite contradictory with what was seen qualitatively. Visually, the FSRCNN performed poorly after x2 unable to maintain the branches with lowest SNR in the generated images and CycleGAN showed good results like RCAN and pix2pix up to x8. This is not reflected in the IQA, which points to the fact that more targeted analysis focusing on the biological features and porosity network connectivity should be included in the SR models assessment.

## 3.3. 2D Connected Component Analysis

A connected component (CC) analysis was performed to identify the morphological similarities between the generated and GT images. The analysis was performed on vesselness-filtered images, binarized using hysteresis thresholding following the procedure described in section 2.5.2. Figure 7 shows CC maps for regions of interest of generated images. Component maps were shown on the same scale based on component area, between 50 and 8000 pixel$^2$. Maps in Figure 7 are therefore best used to compare relative size of large components to the GT. Small components are difficult to identify due to lack of contrast from the background. Maps with a more varied colour map and therefore displaying all components regardless of size are shown in Supplementary Figure 7.

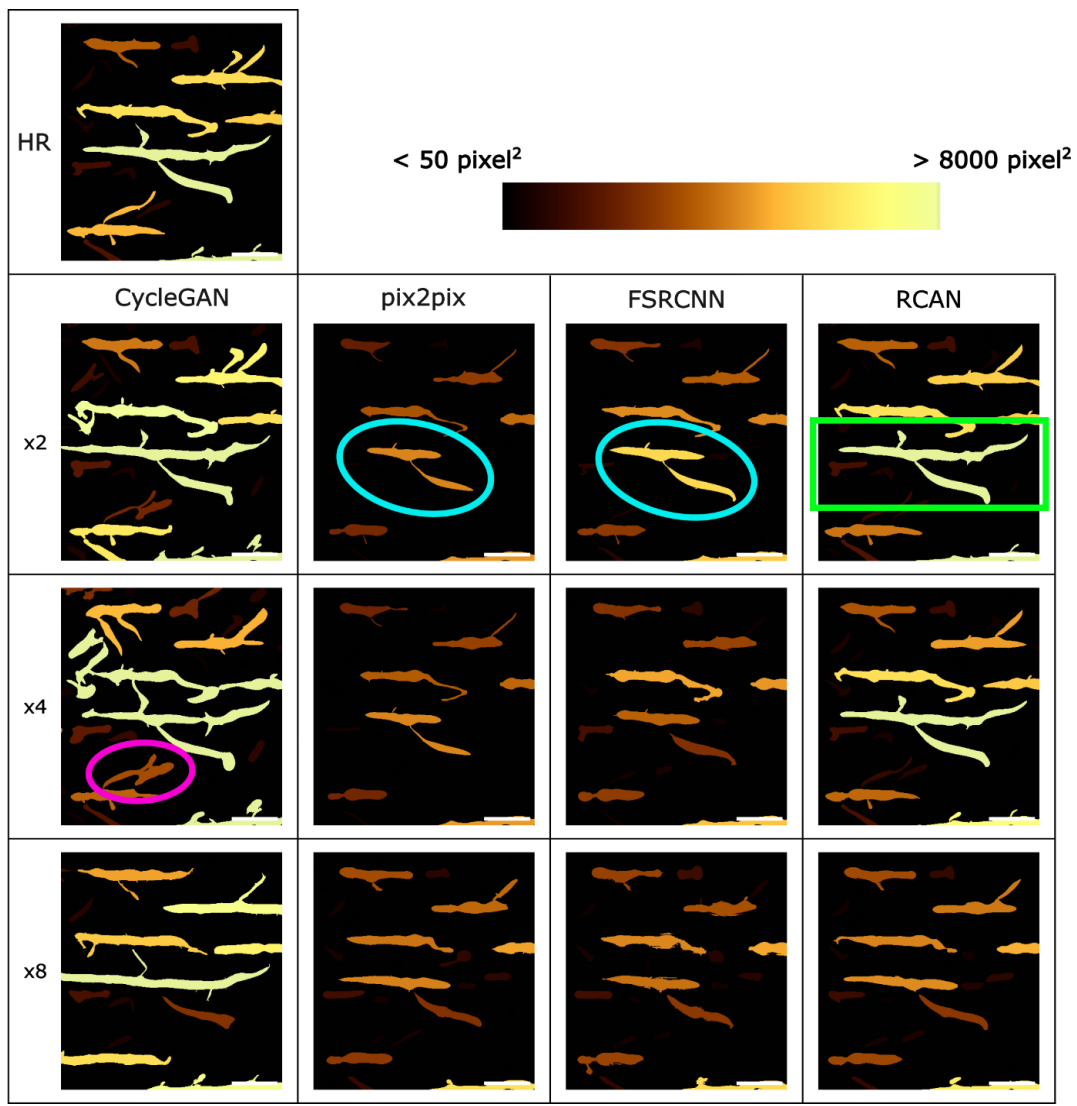

*Figure 7: Connected component maps for a representative 480 × 520 pixels region of interest of generated images at each experimental resolution. The CC map of the GT HR is shown at the top. Components are labelled based on pixel area of each component, and all maps are shown on the same colour scale from 50 pixel$^2$ to 8000 pixel$^{2,}$, and are respectively represented as lightest to darkest colours, disregarding the background. Any components smaller or larger than these values are labelled as their respective extrema value. A pink circle in CycleGAN x4 shows an example of two*

*components that have been merged. Blue circles in pix2pix and FSRCNN x2 show components that are maintained but with a smaller area compared to the GT. A green box in RCAN x2 shows an example of a component that appears very similar to the GT. Note that small components are not visible due to low contrast with the background.*

Qualitatively, the CC maps show that CycleGAN appears to maintain the most similar components to the GT map. CycleGAN enlarges structures, with more components joined and small components appearing larger. An example is shown by a pink circle at CycleGAN x4, where merging was the most evident. CycleGAN, however, generated many false positives at x8, mostly represented by small components. This is visible in Supplementary Figure 7, where colour maps are individualized between models, with no maximum or minimum threshold. The x8 CycleGAN map shows most large components at the same colour scale, which suggests that there are many small components. RCAN appears to perform very similarly to the GT at x2 and x4, with some disconnections but overall, many similar sized components. Performance decreases at x8 with loss of some smaller components and separation of larger ones. Pix2pix and FSRCNN appear to perform similarly at x2, after which FSRCNN performance decreases, no longer maintaining many similar branches to the GT. At x4, pix2pix appears to identify many structures and connections from the GT, but with much smaller areas. This is most evident in Supplementary Figure 7. Pix2pix continues to decrease in performance at x8 with more disconnections and lost components. Overall, CycleGAN appears most consistent with maintaining GT components across all resolutions.

Three important CC metrics are shown in Figure 8, and the remaining can be found in Supplementary Figure 8. Metrics are calculated as a percentage compared to the GT number of CC. The most important values from the CC analysis are the percentage of matching components and the number of missing components. These values show first how many components are exactly generated by the models, and how many were lost. At x2, for matching components, CycleGAN, FSRCNN, and RCAN perform similarly, with no significance between results. Pix2pix shows slightly lower performance than other models, with significant difference from the RCAN. Performance for all methods decreases with resolution, with no significant difference between models at x8. More information can be deduced from the number of missing components, where we see good performance from CycleGAN for all resolutions, continuously outperforming the other models. Pix2pix appears to remain consistent, with a slight increase at x8, however it remains at round 20% missing components compared to GT. Beyond x2, FSRCNN decreases in performance with a great increase in number of missing components. Missing components for RCAN greatly increase at x8. Finally, CycleGAN shows the largest number of false positives, with a great increase between x4 and x8, showing a significant difference from the RCAN. All other metrics have very few false positives until x8, where pix2pix and FSRCNN values increase.

Similarities in matching components between all models does not provide conclusive results on model comparison, with all models performing similarly and with consistent significance between the best and worst performing model. CycleGAN, however, shows a lower instance of missing components and higher number of false positives compared to other models for all resolutions. Missing components and false positives could both have negative impacts on analysis of dentin porosity, either leaving gaps in the porosity network or adding false connections. A trade-off between missing components and false positives could therefore be investigated further, and depending on the nature of these components one could be preferred. For example, as is apparent in Supplementary Figure 7, CycleGAN false positives are for the most part small noise components which could be further filtered and therefore may not pose as negative of an effect on network analysis compared to missing components.

CC analysis performed in 2D reflects the serial image acquisition by confocal microscopy. However, dentin porosity forms a 3D network, and the acquired data represent the sampling of this 3D structure. Hence, further analysis should be conducted to understand which CC features are more or less important on a 3D network scale.

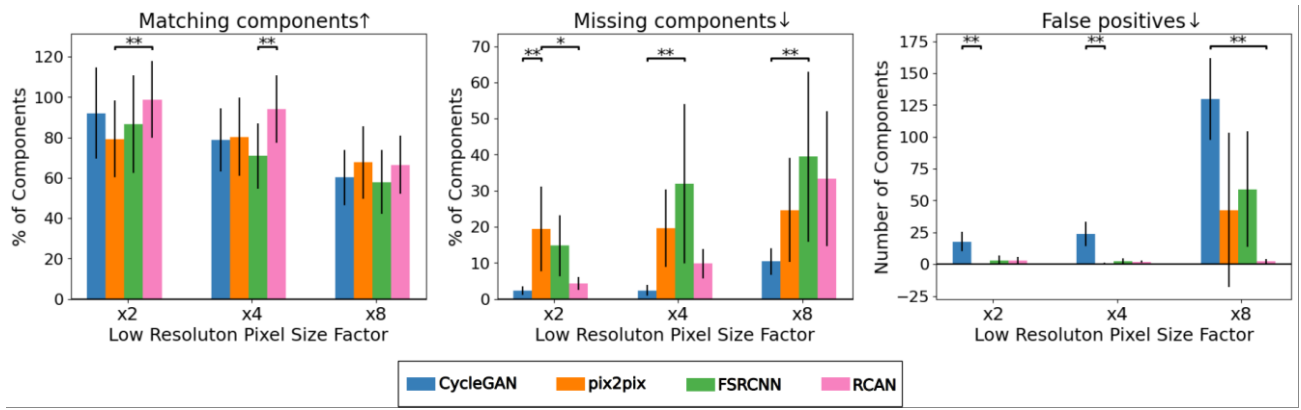

*Figure 8: Matching components, missing components, and false positives for all models and resolutions. Matching and missing components are shown as percentages compared to the number of CC in the GT images, and false positives is measured as a number. Vertical error bars represent the variance for each score measured in the 5 imaged regions. Statistical differences between models are indicated with corresponding p-value when significant. Upward and downwards pointing arrows beside metric names detail whether high scores or low scores represent better results.*

## 3.4 3D Graph connectivity Network Analysis

Confocal fluorescence microscopy allows generating depth-stacks of the cellular porosity, thus allowing to reconstruct a 3D volume whose sampling is generally different in the imaging plane and in depth. In this study, SR modeling is performed directly on the 2D images of the stack but, ultimately, the analysis of the porosity network is conducted in 3D [2]. So, in addition to performing a strict comparison between GT and SR generated 2D images, it is also important to assess how connectivity is preserved in 3D. A graph representation of the binarized images was therefore extracted following the procedure described in section 2.5.2. Figure 9 illustrates the result of this 3D analysis in the form of maximum intensity projections (MIP) along the depth axis of the network skeleton for the GT and SR in a small region of interest. This provides a visual basis for comparison of existing and missing links extracted from the generated images

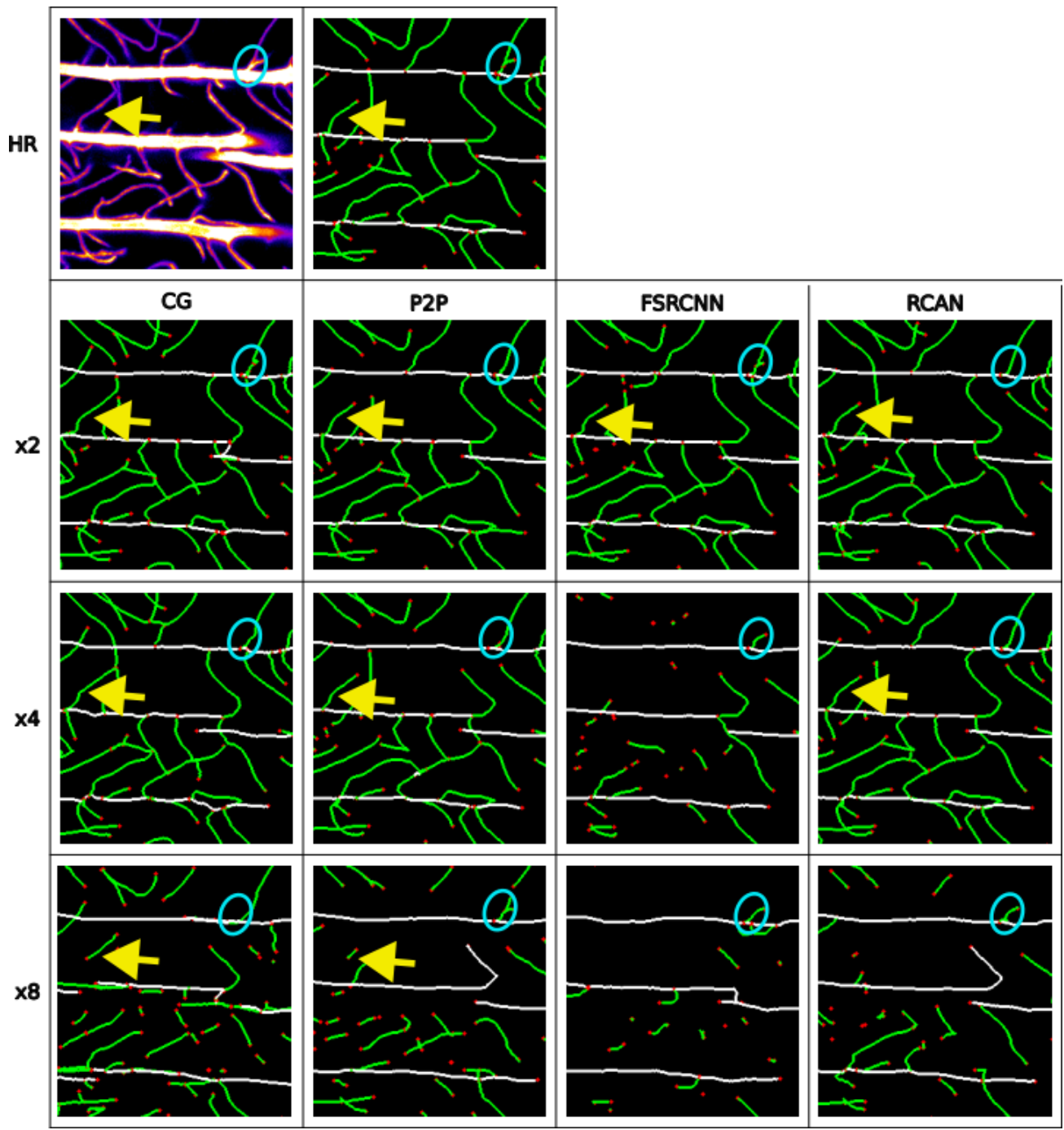

*Figure 9: Graph network extraction for a representative 240 × 260 pixels region of interest of generated images at each experimental resolution. The top row shows GT maximum intensity projection of the original image, with its extracted graph network. White lines represent tubule edges, green lines represent branch edges, and red dots represent nodes. Yellow arrows show an example of a branch that has been generated by certain models and resolutions, but in some cases the connection was lost. A blue circle shows one example of a connection that was maintained by all models, irrespective of whether the entire branch was maintained.*

Visually, tubule edges are well preserved for all models and resolutions, but major differences are observed at the branch level. FSRCNN fails after x2, with an obvious loss of branch edges and connections. CycleGAN appears the closest to the GT, maintaining branch edges and connections at x4, with some edge or node disconnections at x8. In some cases, only part of a branch edge is generated, losing its connection with a tubule, as is illustrated by the yellow arrow at x8. RCAN and pix2pix show similar performance, remaining similar to the GT at x4, however with some disconnections or missing edges. At x8, more information is lost with more missing branch edges and connections.

A more representative analysis was performed on the 3D graphs of the entire image stacks of the 5 measured regions. Six metrics chosen for their representativity are shown in Figure 10, while the remaining calculated metrics are shown in Supplementary Figure 9 and follow similar trends.

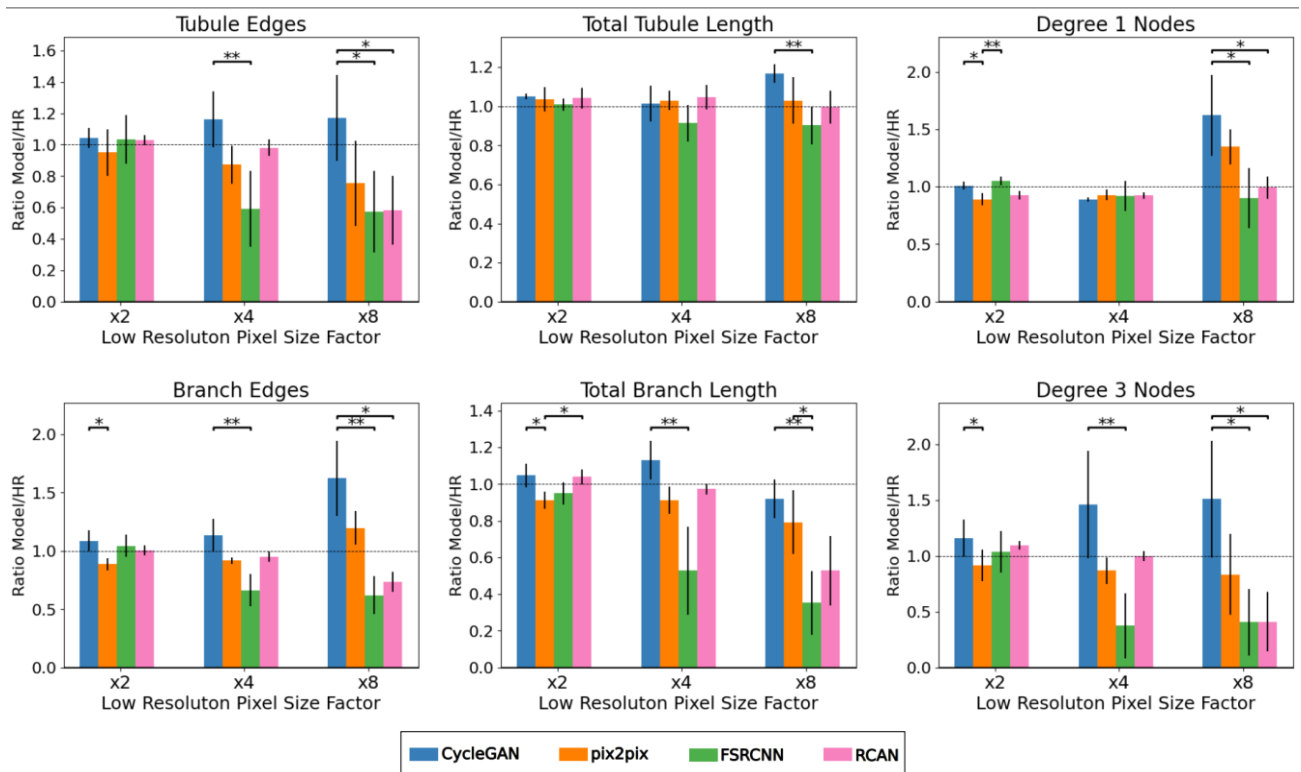

*Figure 10: Graph metric scores for all models and resolutions. The first column shows number of tubule and branch edges, followed by tubules and branch length in the second column, and number of degree 1 and degree 3 nodes in the last column. Scores are calculated as a ratio between the model value and the GT, with a black dashed line showing the optimal score (1). Vertical error bars represent the variance for each score measured in the 5 imaged regions. Statistical differences between models are indicated with corresponding p-value when significant. Upward and downwards pointing arrows beside metric names detail whether high scores or low scores represent better results.*

All studied graph metrics show that CycleGAN and pix2pix are close to the GT without any statistically significant difference. CycleGAN is always higher where pix2pix is lower. RCAN performance is approximately equivalent to pix2pix up to x4 but clearly decreases at x8. FSRCNN does not perform well above x2 and will not be discussed further. More specifically, the number and length of tubule edges is well captured by the three models for all degradation levels with the notable exception of RCAN having the lowest number of tubule edges at x8. So, tubules, which are the largest and brightest features in the images, are well captured overall. Branches, which are the faintest features in the images, with lower SNR are therefore more important. Their corresponding number and length of edges is well preserved up to x4. At x8, CycleGAN seems to have an abnormally high number of edges with respect to pix2pix and RCAN, but their total length is well preserved. This correlates well with the high number of CC false positives of very small dimensions previously noted, which is also reflected in the higher number of degree 1 nodes that typically indicate fragmented branches. Interestingly, the number of degree 3 nodes, which represent a branch stemming from a tubule, or splitting in two and therefore represents a key aspect of the porosity network, appears better preserved with pix2pix, overestimated with CycleGAN and underestimated with RCAN at x8. But, again, the variance between the different regions analyzed is too high to differentiate these three models with a high degree of certainty.

Overall, the results from the 3D graph analysis are quite consistent with those performed with connected components in 2D and also agree well with our initial visual perception. More importantly, all three clearly contradict with what was found from 2D IQA results.

# Discussion

In this study, we investigated the potential of deep learning SR for confocal microscopy imaging of dentinal porosity in teeth. The choice of the tested models was based on recent work by Jhuboo *et al* [18]. As such, the chosen models are not the most recent, nor advanced published SR models, but they are well known and have been benchmarked, which is an advantage for our investigations. Our results show that the question of the reliability of IQA is, indeed, central and must be addressed prior to evaluating SR model performance. Different approaches were tested in this study to compare the quality of image generated at different levels of experimental degradation. The results from the previous sections are complex, but qualitatively summarized in Table 2 and discussed below.

*Table 2: Summary of the different image quality assessment strategies from the different models at all tested resolutions compared to visual perception (VP), in bold.*

|    | CycleGAN | | | | Pix2pix | | | | RCAN | | | | FSRCNN | | | |
|----|----|----|----|----|----|----|----|----|----|----|----|----|----|----|----|----|
|    | **VP** | IQA | CC | GN | **VP** | IQA | CC | GN | **VP** | IQA | CC | GN | **VP** | IQA | CC | GN |
| x2 | **++** | -- | ++ | ++ | **++** | + | + | + | **++** | ++ | ++ | ++ | **++** | + | + | ++ |
| x4 | **++** | -- | ++ | + | **+** | + | + | + | **+** | + | + | ++ | **-** | + | - | -- |
| x8 | **+** | -- | + | + | **-** | + | - | + | **-** | + | - | - | **--** | - | -- | -- |

We first established a classification based on expert visual perception focusing on the topological features of the cellular porosity network. This involved, in particular, our capacity to recognize multiscale porosities, tubules and branches and their connectivity. This revealed obvious differences between SR models, with FSRCNN rapidly failing, contrary to other models performing relatively well up to x8. Subtler differences can also be noted, with CycleGAN efficiently capturing less intense branches as compared to pix2pix and RCAN, but seemingly enlarging the image features. As can be seen in Figure 1 and 5, x8 corresponds to an experimental under-sampling condition (4x the PSF size) where many branches are barely distinguishable by eye, which therefore represents a challenge for image restoration. However, this analysis remains qualitative and biased by our visual perception, which strongly depends on the image contrast of the weakest branches.

While the development of IQA remains an important topic of research for DL applications, the use of standard IQA remains prominent [8]. A selection of full-reference and distribution-based metrics were used in this work to compare model performance which, overall, yielded contradictory results to visual perception. In particular, CycleGAN, which we observed to correctly preserve biological features, was consistently rated significantly worse than other models. FSRCNN, which clearly failed from an early stage was often not very different from pix2pix and RCAN which appeared very close. This is reflected in particular for PSNR, MSE and all SSIM-derived metrics (MS-SSIM and IW-SSIM) in Figure 6 and Supplementary Figure 6. This means that even for such apparently simple data containing 3 classes of features: background, small vessels (branches) and larger, brighter ones (tubules), metrics relying on sole contrast (PSNR) or also including structure (SSIM-type) fail. Furthermore, these metrics are pixel-based, and have therefore been reported to not properly capture perceptual quality of images [52]. Other metrics such as HaarPSI, WD, FID or KID seemed better at estimating the difference between models, except for CycleGAN, which was systematically under-rated. If we exclude CycleGAN, it is interesting to note that HaarPSI is the closest full-reference metric to visual perception. HaarPSI is based on the strongest multiscale wavelet coefficients, making it less sensitive to high frequency intensity fluctuations. Initial comparison of HaarPSI to previous metrics (PSNR, SSIM) showed consistent significantly better correlation to human perception, which matches our findings [33]. This suggests that non-linearities in SR image generation strongly limit other simpler full-reference methods. Similar conclusions can be drawn for WD, FID, KID which assess statistical distance between image intensity distributions and are therefore less prone to minor local fluctuations. Altogether, CycleGAN was found to exhibit the strongest non-linearities of all models. For example, Figure 5 shows that the small branches indicated with green arrows appear at a similar intensity level than larger more intense tubules in the generated images, while other models maintain the intensity difference between the two features. More problematic, the average intensity level was found to strongly fluctuate between patches as observed in Supplementary Fig.4. As demonstrated, this can be corrected, but ultimately, such fluctuations logically impact all IQA metrics.

To better evaluate IQA metrics and quantify model performance, a biology-driven analysis of the porosity network was conducted. First, using connected components calculated independently for each acquired 2D image, i.e. without considering any relation between images of the stack, and second by performing 3D graph analysis of the image stack volumes. Both analysis yielded much closer results to our visual perception than IQA metrics, as summarized in Table 2. Moreover, in addition to better evaluating model performance, this strategy allowed identifying the nature of the differences between GT and generated images. In particular, the results of the CC analysis in Figure 8 and Supplementary Figure 8 show that all models generate a similar amount of matching components (within the statistical error range), which gradually decreases from ~90% at x2 to ~60% at x8. The most important differences between models therefore appear to be the missing components and false positives.

CycleGAN systematically had the smallest fraction of missing components, remaining below 10% and significantly outperforming all other models. This is a strong reason for our initial qualitative performance estimation: tubules and branches appeared to be better retained than with other models. To better understand SR model behaviours, the missing CC were identified in the GT images and their corresponding intensities distribution was overlaid on the global histogram distribution of the GT images as illustrated for pix2pix in Figure 11. This allows concluding that missing components in pix2pix, RCAN and FSRCNN are systematically the smaller branches with lowest intensity pixel values, despite our efforts to balance classes in the training dataset.

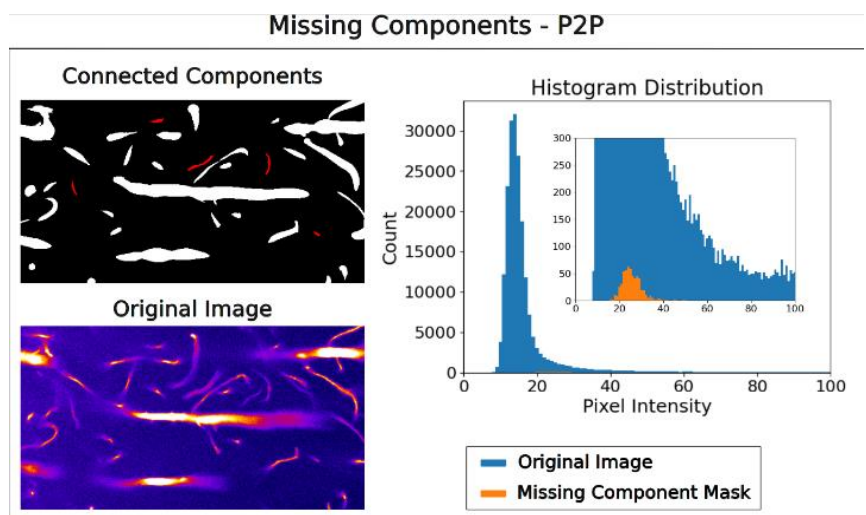

*Figure 11: Example of missing components from the pix2pix model. Missing components are shown in red with modeled components shown in white. Histogram distributions were calculated on original image masked with missing connected component mask, showing in orange the pixel values from the GT image that were missed during connected component analysis.*

The situation is reversed when considering false positives, which are relatively scarce up to x4 for all models but CycleGAN and even < 5% with RCAN at x8. Again, it is important to understand what the false positives correspond to in the GT images. The analysis summarized in Figure 12 shows two different types of false positives found in the CycleGAN. First, some false positive CCs corresponded to very faint branches that were not found in the GT but were actually present in our HR images as seen in Figure 12.a. This highlights a limitation of the CC analysis, which is based on a complex segmentation pipeline that inevitably misses some extremely faint features. Since CycleGAN tends to increase intensities of branches and tubules to similar intensity levels, those very weak branches in HR images are enhanced by CycleGAN and picked up by subsequent segmentation. So, in some sense, this suggests that CycleGAN is beneficial for our standard analytical segmentation protocol, acting as an enhancement filter for generated features. This is somewhat mitigated by the fact that a much larger fraction of CycleGAN false positives CCs actually correspond to noise, as observed in Figure 12.b. This is related to the high sensitivity of the model, but in this case, it has a negative impact for SR. However, those very small components could easily be removed or filtered out by size, if necessary, and are therefore not considered very problematic.

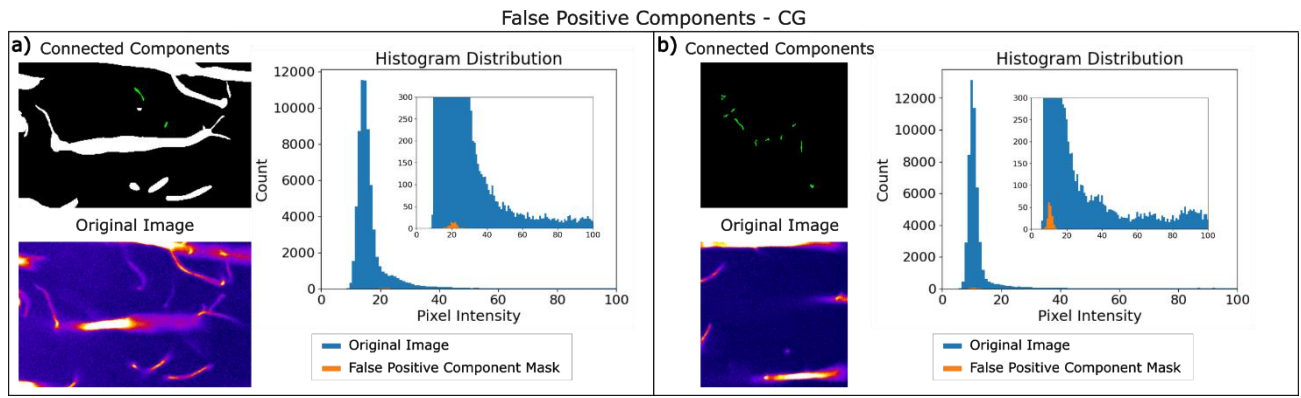

*Figure 12: Two examples of false positive connected components compared to the original GT image. False positive components are shown in green, with white showing components that were generated by the model. Histogram distributions were calculated on the original GT image masked with the false positive components, showing in orange values from the original image that were labelled as a connected component. a) Shows a case where false positives show very low intensity branches, not identified in the GT CC analysis. b) Shows a case where false positives show noise.*

It was observed that low intensity branches were missing, showing that pix2pix struggled to reconstruct smaller branches with low intensity pixel values. Similar observations were seen with RCAN. As discussed above, it is important to note that the CC analysis performed with our segmentation pipeline is very sensitive to the non-linear behaviour of the SR models. This reflects in both false positive and missing components, but also in split and merged components of SR, as visible in Figure 7 and quantified in Supplementary Figure 8. So overall, CC analysis provides new mechanistic insight into model performance, with results more in line with our visual perception of the biological features of the network.

The final proposed assessment focuses on the 3D connectivity of biological features visualized throughout the image stack. Fluid-flow in porosity presents the main mechanosensing theory of dentin [3], [53], [54], so it is essential that SR models achieve good performance in preserving tubule and branch connections. Our graph results unambiguously showed that the largest, most intense tubules were globally well retrieved by all models at all resolutions. The major differences were observed on branch edges and their connectivity to the main tubules, which decrease as a function of resolution. However, Figure 10 unambiguously showed that pix2pix was closest to GT and most stable at all resolutions, although slightly under-performing by less than 10%. The difference with RCAN is more pronounced than with all previous analysis. As expected from the CC analysis, CycleGAN also performs very well, but over-performs at x8. This was clearly expected from previous observations: this model enlarges all features and, as a result, it merges components.

Of course, assessing the efficiency of a 2D model in 3D is somewhat far-stretched, and one could expect that 3D models would perform better to maintain 3D structures, as shown in recent studies [55], [56]. However, our results demonstrate that pix2pix and CycleGAN already perform surprisingly well in preserving 3D connectivity. Since 3D models require much more computational power, using lighter 2D models can be seen as an advantage. Furthermore, memory handling of 3D inputs might require decreasing the input data volumes beyond what is reasonable with respect to our feature sizes. Not to mention that confocal microscopy is often very limited in depth. So the adaptation of a 2D model to 3D is far from straightforward.

Globally, our results tend to show that standard IQA metrics are not well adapted for assessment of SR performance in preserving overall dentin porosity features in 2D, in particular small branches and, as a consequence, in retrieving 3D network connectivity. Interestingly, one would have expected metrics strongly influenced by structure, such as SSIM-types, to perform better, which is not the case. Other studies have drawn similar conclusions, with SSIM unable to reliably quantify perceptual quality [8], [57], [58]. This is even more surprising with the geometry score (GS), which defines quality on the basis of pure image topology [37]. This metric was found to exhibit a large variance across samples which prevented reliable comparison. However, as for other distribution-based metrics, the current implementation of GS calculates feature vectors of the images extracted by a pre-trained InceptionV3 net. This classifier was pre-trained using labeled images from ImageNet, a database containing millions or real images [39], [59]. So, in addition to potential weaknesses in the reduction of the persistent homology analysis, it could very well be that this classifier remains limited for our biological settings. Retraining the InceptionNet with our data might solve this issue but would require extensive data labeling, therefore remaining very time consuming. This is a general problem in the development of new IQA metrics involving DL recognition of content: large image amounts of labelled data are required to train models which are computationally expensive and their generalizability is questionable [12]. All those limitations probably explain why standard IQA is still being heavily used. Nevertheless, our structural analysis of dentinal porosity shows that the ingredients to successful and maybe more simple SR-IQA of dentin porosity could be: morphological filtering based on Hessian and connected components or graph reduction.

In all cases, the ability to perform a full-reference IQA assessment is limited by the availability of paired data. Our unique experimentally acquired HR and LR paired dataset was imperative not only for paired model training, but also for the assessment of all model performance, including unpaired CycleGAN. Without experimentally paired data, synthetic data generation would be required which would only be as good as the physical and biological model used to create a realistic synthetic dataset. Mineralized tissues are very complex media and are not well characterized optically so far, so such a task is absolutely non-trivial. From an experimentalist point of view, acquiring paired data remains much less complex and the data acquisition strategy described in this study might prove valuable for similar studies of other biological systems. Our dataset could therefore be used both for training/testing new models, as well as testing new IQA metrics.

The models tested in this study also represent very different classes: supervised (pix2pix, RCAN, FSRCNN) or unsupervised (CycleGAN), and very different architectures including a CNN (FSRCNN), an attention network (RCAN), and two GANs (pix2pix and CycleGAN). The following observations can be made on the basis of our CC and graph analysis:

- The rapid failure of FSRCNN beyond x2 could be anticipated since SR-CNNs have previously been found to perform poorly with large scaling down-sampled factors [6]. However, x2 corresponds to the instrument PSF size and actually represents a common imaging parameter. So this model is therefore of very limited practical interest beyond computing speed.

- RCAN is a very deep residual channel network that uses residual-in-residual structure to improve flow of low-frequency information using both long and short skip-connections that was previously found to perform better for SR tasks [21]. Skip connections and channel attention components of this network aim to put more focus on important features, which should ideally show better performance compared to more simplistic models. It showed good performance up to x4, which represents an image under-sampling by a factor 2. This already provides a substantial improvement of scan time by a factor of 10. Note that this model was trained with an L1 loss, which has been reported to encourage blurry results and miss high-frequency details in images [20]. The fact that RCAN relies solely on this loss could hinder performance compared to pix2pix and CycleGAN, that incorporate other loss functions.

- CycleGAN and pix2pix performed better than the others up to x8, which was considered a high level of image degradation. Both are conditional GANs built with very similar architecture: the CycleGAN was adapted from the pix2pix to create an unpaired method [19],[20]. The discriminator used in both cases is a patchGAN, which penalizes structure on 70x70 pixel image patches as opposed to a global value, which could be beneficial to focus on small branches in the image that may be less weighted in a global score [20]. Future considerations could evaluate if different patch sizes could further improve pix2pix and CycleGAN performance. pix2pix also benefits from a U-Net architecture with skip connections which could help maintain important information such as prominent edges [20]. A common struggle with GANs is the difficulty to train stable models [60], as illustrated by the instabilities shown in Supplementary Figure 3. Furthermore, for CycleGAN, the model trained with x4 and x8 data showed mode failure approximately halfway through training. However, even with unstable training and a shorter trained model, good performance was achieved, showing great promise for these GAN models which often considered outdated, but have room for better performance [52]. Note that using an unpaired model greatly simplifies data acquisition in cases where acquiring data in both domains proves challenging, for example across multiple modalities.

In this study, we could achieve a reduction in scan time by a factor 20.3 using CycleGAN or pix2pix. This already represents a considerable improvement for dental research application since we could expect that a full tooth section could be mapped in approximately 90 hours. This could certainly be improved with better tuning of those models or with more recent ones. But it would already represent a considerable achievement and open new avenues to better understand dentin formation in healthy and pathological cases.

Some limitations of this study include the use of models not optimally tuned for this dataset, leading to failure in training for the x4 and x8 CycleGAN model, and potentially decreasing performance of all models. However, keeping parameters according to previous publications was consistent for all models and therefore no bias between models was introduced. In future work, hyperparameter tuning for different models could be done to ensure best possible performance for each specific dataset. Secondly, data was acquired from only one tooth for training, and all data were acquired within a 300 µm distance from the DEJ, a structurally specific tooth region. Such that the trained models most likely lack generalizability to different teeth, or different regions of the tooth where branch and tubule orientation or distribution could be different. To try mitigating the effects of different tubule orientation, rotation was used as a data augmentation method. Furthermore, a different sample was used to assess model performance to show that models could perform well on data that were not included in the training set. In future studies, more samples and a wide variety of regions of dentin should be used to improve generalizability of the models. A final limitation was the use of 2D models on 3D datasets. Since the biological structure is inherently tridimensional, 3D models should be preferred. However, those are often computationally complex and are also challenging experimentally. Nevertheless, the tested 2D models proved remarkably efficient in preserving 3D connectivity, that should be compared in the future with the results from 3D models.

# Conclusions

The present study investigated the use of four DL models for SR with a unique set of experimentally acquired pairs of LR/HR confocal fluorescence microscopy images of dentin porosity. Standard IQA metrics used to compare model performance were found to be in contradiction with visual perception. An in-depth analysis focusing on the distinct multiscale nature and of the biological porosity network was therefore conducted using connected components and graph network analysis. The quantification of 2D morphological and 3D connectivity thus provided a biologically relevant assessment of generated SR images. This approach showed that CycleGAN (unsupervised) and pix2pix (supervised), maintained branches and connections better than other models and with similar performances up to x8, RCAN up to x4, and FSRCNN up to x2 pixel size. The use of biology-driven assessment of SR results provided a clearer mechanistic investigation of model performance. For instance, it allowed identifying missing components and false positives generated by SR, which mostly originate from the specific model sensitivity to very weak intensity features. This also highlighted the importance of non-linearity found in all models, but particularly important with CycleGAN which ultimately explains the failure of IQA metrics. From a practical point of view, confocal acquisitions at x8 pixel size correspond to an under-sampling up to a factor 4 with respect to the microscope PSF, and the resulting image degradation represents a considerable challenge for image restoration. Furthermore, this speeds up acquisitions by a factor of 20.3 which represents a considerable improvement which opens the way for porosity imaging at the whole tooth scale, an important challenge for fundamental dental research.

# Ethics Statement

# CRediT AUTHOR STATEMENT

Lauren Anderson: Conceptualization, Data curation, Formal analysis, Investigation, Software, Visualization, Writing – original draft
Lucas Chatelain: Investigation, Software
Nicolas Tremblay: Methodology, Supervision, Writing - Review & Editing
Kathryn Grandfield: Funding acquisition, Methodology, Supervision, Writing - Review & Editing
David Rousseau: Conceptualization, Methodology , Supervision, Writing - Review & Editing
Aurélien Gourrier: Conceptualization, Funding acquisition, Methodology, Project administration, Resources, Software, Supervision,  Writing - Review & Editing

# Acknowledgements

We thank Elsa Vennat (Université Paris-Saclay, CentraleSupélec, ENS Paris-Saclay, CNRS, LMPS-Laboratoire de Mécanique Paris-Saclay, Gif-sur-Yvette, France) and Elisabeth Dursun (Université Paris Cité, U1333 Santé Orale, INSERM, Dental School, Montrouge, France & AP-HP, Service de médecine bucco-dentaire,Hôpitaux Universitaires Henri Mondor, Créteil, France) for providing tooth samples, and Seunghwan Goldmund Lee (Univ. Grenoble Alpes, CNRS, LIPhy, Grenoble, France & Pusan National University, Department of Opto & Cogno Mechatronics Engineering, Busan, Republic of Korea) for sample preparation assistance. Funding for this work was received from the Human Frontier Science Program (HFSP, https://www.hfsp.org/) Research Grant RGP0023/2021 and from the French National Research Agency program France 2030, grant ANR-23-IACL-0006.

# Data Availability

All confocal fluorescence microscopy images used to create the training dataset are publicly available from Zenodo: https://doi.org/10.5281/zenodo.17185790 under Creative Commons Attribution 4.0 International licence. The Python code used to analyze the data and produce the results is available without restriction under the BSD-3 licence conditions, on the following Gitlab repository of the University Grenoble Alpes: https://gricad-gitlab.univ-grenoble-alpes.fr/mintissliphy/studentscodes/pysrmintiss.git (commit 3f4b109926c33060cfc19eb00387caa3bb52ee20)